\documentclass[
]{ceurart}

\usepackage{listings}
\usepackage{algorithm}
\usepackage{algpseudocode}
\usepackage{todonotes}
\usepackage{soul}
\usepackage{multirow}

\begin{document}

\copyrightyear{2025}
\copyrightclause{Copyright for this paper by its authors.
  Use permitted under Creative Commons License Attribution 4.0
  International (CC BY 4.0).}

\conference{CLEF 2025 Working Notes, 9 -- 12 September 2025, Madrid, Spain}

\title{TurQUaz at CheckThat! 2025: Debating Large Language Models for Scientific Web Discourse Detection}

\title[mode=sub]{Notebook for the CheckThat! Lab at CLEF 2025}

\author[1]{Tarık Saraç}[%
orcid=0009-0004-7598-5825,
email=tariksarac@etu.edu.tr
]
\fnmark[1]

\author[1]{Selin Mergen}[%
orcid=0009-0002-3284-9490,
email=s.mergen@etu.edu.tr
]
\fnmark[1]
\cormark[1]

\author[2]{Mucahid Kutlu}[%
orcid=0000-0002-5660-4992,
email=mucahidkutlu@qu.edu.qa
]
\cormark[1]

\address[1]{TOBB University of Economics and Technology,
Ankara, Türkiye}
\address[2]{Qatar University, Doha, Qatar}

\cortext[1]{Corresponding author.}
\fntext[1]{These authors contributed equally.}

\begin{abstract}
    In this paper, we present our work developed for the scientific web discourse detection task (Task 4a) of CheckThat! 2025. 
    We propose a novel council debate method that simulates structured academic discussions among multiple large language models (LLMs) to identify whether a given tweet contains (i) a scientific claim, (ii) a reference to a scientific study, or (iii) mentions of scientific entities. We explore three debating methods: i) single debate, where two LLMs argue for opposing positions while a third acts as a judge; ii) team debate, in which multiple models collaborate within each side of the debate; and iii) council debate, where multiple expert models deliberate together to reach a consensus, moderated by a chairperson model. We choose council debate as our primary model as it outperforms others in the development test set. 
    Although our proposed method did not rank highly for identifying scientific claims (8th out of 10) or mentions of scientific entities (9th out of 10), it ranked first in detecting references to scientific studies.
    \end{abstract}

\begin{keywords}
    Scientific Discourse Detection \sep
    Debating Method \sep
    Large Language Models
\end{keywords}

\maketitle




\section{Introduction}
In this work, we present our approach for Subtask 4a (Scientific Web Discourse Detection) \cite{clef-checkthat:2025:task4} of the CheckThat! 2025 shared task \cite{checkthat}. The subtask involves three binary classification problems for a given tweet:

\begin{itemize}
    \item \textbf{Category 1:} Does the tweet contain a scientific claim?
    \item \textbf{Category 2:} Does the tweet reference a scientific study or publication?
    \item \textbf{Category 3:} Does the tweet mention scientific entities, such as a university or scientist?
\end{itemize}

In this work, we propose a few-shot classification approach in which LLMs engage in debate to reach a final decision. Specifically, we introduce three distinct debating strategies: \textit{single debate}, \textit{team debate}, and \textit{council debate}. In the single debate setting, two LLMs argue from opposing perspectives, while a third model serves as the judge. In team debate, multiple models collaborate on each side; team members first discuss internally before presenting their collective arguments to the opposing team. Finally, in the council debate approach, a group of expert models discuss together to reach a consensus, moderated by a chairperson model. As we observed a strong correlation between the positive classes in Category 2 and Category 3, we adopt a simple heuristic: if our method predicts that a tweet contains a reference to a scientific study or publication (Category 2), we also label it as positive for Category 3 (i.e., containing mentions of scientific entities). 

In our experiments, we utilize several LLMs, including Gemma3 (12B) \cite{gemma3}, Qwen3 (8B) \cite{qwen3}, DeepSeek-R1 (8B) \cite{deepseekr1}, Phi-4 (14B) \cite{phi4}, Mistral (7B) \cite{mistral}, LLaMA 3.1 (8B) \cite{llama3}, o4-mini\footnote{\href{https://openai.com/index/introducing-o3-and-o4-mini/}{https://openai.com/index/introducing-o3-and-o4-mini/}}, and Claude-4\footnote{\href{https://www.anthropic.com/news/claude-4}{https://www.anthropic.com/news/claude-4}}. When using only open-source models, the council debate method outperforms  other approaches on the development set, and the team debate method is superior to the single debate method. Based on these results, we selected the council debate method as our primary method.

In the official rankings, our model ranked $8^{th}$ ($F_1 = 0.7273$) in Category 1 (detecting scientific claims) and $9^{th}$ ($F_1 = 0.7766$) in Category 3 (detecting mentions of scientific entities). However, it achieved $1^{st}$ place ($F_1 = 0.7805$) in Category 2 (detecting references to scientific studies).

\section{Related Work}

\subsection{Scientific Web Discourse Detection}

Detecting scientific claims and related content in social media, especially on platforms like Twitter\footnote{\href{https://x.com}{https://x.com}}, has become a growing area of research in recent years~\cite{clef-checkthat:2025:task4,checkthat,scitweets}. Shared tasks such as CLEF CheckThat! (Task 4a)~\cite{clef-checkthat:2025:task4}, SemEval-2023 Task 8~\cite{semeval2023}, and the FIRE-2023 CLAIMSCAN challenge~\cite{claimscan2023} have focused on distinguishing scientific claims, references, and science-related entities from general online discourse. Most competitive systems in these tasks use supervised learning, with large pre-trained transformer models like BERT or RoBERTa fine-tuned on carefully annotated datasets, often augmented with auxiliary features such as tweet metadata or ensemble techniques~\cite{survey_claim}. While these methods deliver strong results, they rely on large labeled datasets and task-specific training, which can limit adaptability to new domains or languages. In contrast, our approach utilizes the few-shot reasoning capabilities of LLMs and uses a debate-style framework, where multiple models collectively reason and reach a consensus on the presence of scientific discourse, rather than a single model making a prediction. Thus, our methods offer better flexibility and reduce reliance on  annotated large training datasets.

\subsection{Debating LLMs for Scientific Discourse Detection}

Recently, debate-based multi-agent frameworks using LLMs have been proposed to improve model reasoning, evaluation, and decision-making~\cite{chateval,debatrix,khan2024debate}. In these systems, multiple LLMs (sometimes with different roles or perspectives) discuss, argue, or collaborate on a task before producing a final judgment, often resulting in outcomes that align more closely with human assessment than those from a single model. Earlier debate methods typically featured two agents (one "pro" and one "con") with a judge, or small fixed panels, mainly for open-ended generation or evaluation tasks~\cite{wang2023chatgpt,taubenfeld2024debate,estornell2024multi}. However, homogeneous panels can sometimes fall into biased agreement or fail to explore diverse viewpoints~\cite{estornell2024multi}. Eo et al. \cite{eo2024down} address efficiency by triggering debates only on uncertain cases. Our council debate method builds on these ideas by using a more diverse set of LLM agents, with each agent contributing its own reasoning. Unlike previous works focused on generation quality evaluation, we apply the debate framework directly to scientific discourse classification in social media. By combining several LLMs in a structured decision process, our approach aims to reduce individual model bias and promote more careful evaluation, resulting in more reliable detection of scientific claims, references, and entities in noisy online environments.

\section{Proposed Methods}

\begin{figure}[ht]
    \centering
    \includegraphics[width=\textwidth]{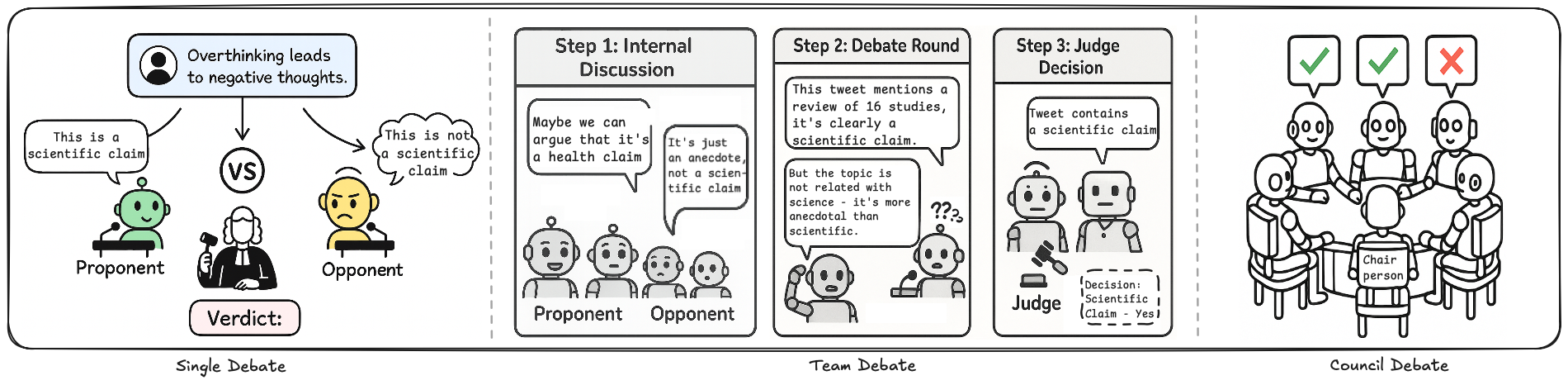}
    \caption{Overview of our three debate frameworks: Single Debate, Team Debate, and Council Debate. Each method involves structured interactions among large language models to reach a final decision.}
    \label{fig:debate_framework}
\end{figure}

We explored three different multi-LLM approaches for scientific discourse detection, each building upon the previous method's insights. All approaches use structured deliberation processes but differ in their organizational structure and decision-making mechanisms. In our methods, we use several different prompts. The actual prompts used in our study are given in Appendix \ref{appendix:prompts}. 
We apply each method independently for each category. However, when a tweet references a scientific study (Category 2), we also assign a positive label to Category 3 (i.e., containing scientific entities), as such references typically imply the presence of scientific terms. 
An overview of these three debate frameworks is illustrated in \textbf{Figure~\ref{fig:debate_framework}}.

Now we explain each debate method in detail. 

\subsection{Single Debate Method}

Our first approach implements a traditional debate format where two LLMs argue for opposing positions while a third model acts as a judge. This method processes each category independently through separate debates. 

\textbf{Algorithm \ref{alg_single_debate}} describes our single debate method. For each interaction with LLMs, a brief prompt is given just to explain the prompt. Actual prompts are provided in Appendix \ref{appendix_single}. 
The algorithm takes the tweet and the classification category as parameters. We first set our models used in the debate process and assign specific models to different roles: a proponent model $P$ that argues the tweet contains the category, an opponent model $O$ that argues against it, and a judge model $J$ that makes the final decision [\textbf{Lines 2-4}]. The debate begins with opening statements where the proponent generates supporting arguments and the opponent generates opposing arguments for the given tweet and category [\textbf{Lines 5-6}]. We maintain a transcript that records all arguments throughout the debate [\textbf{Line 7}]. The core debate consists of $R$ rounds where the proponent rebuts the opponent's argument, and the opponent rebuts the proponent's argument in response [\textbf{Lines 8-12}]. After the $R$ rebuttal rounds, both sides provide closing statements to summarize their positions [\textbf{Lines 13-15}]. Finally, the judge model evaluates the complete debate transcript and makes a classification decision for the current category [\textbf{Lines 16-17}]. 

\begin{algorithm}[h]
\caption{Single Debate Method}
\begin{algorithmic}[1]
\State \textbf{Input:} Tweet text $T$, category $c$, maximum rounds $R$
    \State $P \leftarrow$ proponent model for category $c$
    \State $O \leftarrow$ opponent model for category $c$ 
    \State $J \leftarrow$ judge model
    \State $RP_0 \leftarrow$ Generate($P$, "argue tweet $T$ contains category $c$")
    \State $RO_0 \leftarrow$ Generate($O$, "argue tweet $T$ does NOT contain category $c$")
    \State $transcript \leftarrow [RP_0, RO_0]$
    \For{$r = 1$ \textbf{to} $R$}
        \State $RP_r \leftarrow$ Generate($P$, "rebut opponent argument $RO_{r-1}$")
        \State $RO_r \leftarrow$ Generate($O$, "rebut proponent argument $RP_r$")
        \State $transcript \leftarrow transcript + [RP_r, RO_r]$
    \EndFor
    \State $RP_{final} \leftarrow$ Generate($P$, "provide closing statement")
    \State $RO_{final} \leftarrow$ Generate($O$, "provide closing statement")
    \State $transcript \leftarrow transcript + [RP_{final}, RO_{final}]$
    \State $decision \leftarrow$ Generate($J$, "evaluate debate $transcript$ for category $c$")
    \State return ExtractClassification($decision$)
\end{algorithmic}
\label{alg_single_debate}
\end{algorithm}

\subsection{Team Debate Method}

Building on the single debate approach, we propose a team-based method where multiple models collaborate within each side of the debate. This approach maintains the adversarial structure while adding diverse perspectives within each team. Importantly, team members discuss among themselves before presenting their arguments to the opposing team.

\textbf{Algorithm \ref{alg_team_debate}} describes our team debate method. For each interaction with LLMs, a brief prompt is given just to explain the prompt. Actual prompts are provided in Appendix \ref{appendix_team}. 
We assign teams of models to different roles: a proponent team $PT$ that argues the tweet contains the category, an opponent team $OT$ that argues against it, and a judge model $J$ that makes the final decision [\textbf{Lines 2-4}]. The debate begins with internal team discussions where proponent team members collaborate to develop their strategy and arguments, followed by opponent team members doing the same [\textbf{Lines 5-6}]. Each team then presents their opening statements, with all proponent team members arguing for the category and all opponent team members arguing against it [\textbf{Lines 7-8}]. We maintain a transcript that records all arguments throughout the debate [\textbf{Line 9}]. The core debate consists of $R$ rounds where teams first conduct internal discussions to plan their rebuttals, then present coordinated responses to the opposing team's arguments [\textbf{Lines 10-15}]. Finally, the judge model evaluates the complete team debate transcript and makes a classification decision for the current category [\textbf{Lines 17-18}]. 


\begin{algorithm}[h]
\caption{Team Debate Method}
\begin{algorithmic}[1]
\State \textbf{Input:} Tweet text $T$, category $c$, team size $S$, maximum rounds $R$
    \State $PT \leftarrow$ proponent team of size $S$ for category $c$
    \State $OT \leftarrow$ opponent team of size $S$ for category $c$
    \State $J \leftarrow$ judge model
    \State $PT_{strategy} \leftarrow$ InternalDiscussion($PT$, "develop strategy for category $c$")
    \State $OT_{strategy} \leftarrow$ InternalDiscussion($OT$, "develop strategy against category $c$")
    \State $PT_{opening} \leftarrow$ TeamArguments($PT$, "argue tweet $T$ contains category $c$")
    \State $OT_{opening} \leftarrow$ TeamArguments($OT$, "argue tweet $T$ does NOT contain category $c$")
    \State $transcript \leftarrow [PT_{opening}, OT_{opening}]$
    \For{$r = 1$ \textbf{to} $R$}
        \State $PT_{rebuttal\_plan} \leftarrow$ InternalDiscussion($PT$, "plan rebuttals to $OT$ arguments")
        \State $OT_{rebuttal\_plan} \leftarrow$ InternalDiscussion($OT$, "plan rebuttals to $PT$ arguments")
        \State $PT_{rebuttals} \leftarrow$ TeamArguments($PT$, "present coordinated rebuttals")
        \State $OT_{rebuttals} \leftarrow$ TeamArguments($OT$, "present coordinated rebuttals")
        \State $transcript \leftarrow transcript + [PT_{rebuttals}, OT_{rebuttals}]$
    \EndFor
    \State $decision \leftarrow$ Generate($J$, "evaluate team debate $transcript$ for category $c$")
    \State return ExtractClassification($decision$)
\end{algorithmic}
\label{alg_team_debate}
\end{algorithm}

\subsection{Council Debate Method}

Our final approach implements a collaborative council structure where multiple expert models deliberate together to reach consensus, moderated by a chairperson model. This method moves away from adversarial debate toward collaborative decision-making.

\textbf{Algorithm \ref{alg_council_debate}} describes our council debate method. For each interaction with LLMs, a brief prompt is given just to explain the prompt. Actual prompts are provided in Appendix \ref{appendix_council}. 
Firstly, we assign a council of expert models $CM$ and a chairperson model $C$ to moderate the discussion [\textbf{Lines 2-3}]. The process begins with each council member providing an initial assessment and vote for the current category [\textbf{Line 5}]. We then check if the initial votes have reached the consensus threshold $\theta$. If consensus is achieved, we finalize the decision using the majority vote [\textbf{Lines 5-7}]. Otherwise, we proceed with $R$ rounds of structured discussion where the chairperson summarizes the current state and guides the focus, followed by each council member responding to the discussion and updating their position [\textbf{Lines 8-15}]. After each round, we check whether consensus has been reached or if the votes have stabilized to finish the discussion early [\textbf{Lines 12-14}]. 
Finally, we use the majority vote from the final round as our decision [\textbf{Line 16}]. 

To provide concrete examples of how our debating methods work in practice, we present conversation logs from each method in Appendix~\ref{appendix:conversation_logs}, 
 showing how different models interact to reach a final decision.

\begin{algorithm}[h]
\caption{Council Debate Method}
\begin{algorithmic}[1]
\State \textbf{Input:} Tweet text $T$, Category $c$, consensus threshold $\theta$, maximum rounds $R$
    \State $council \leftarrow$ assign council members from $CM$ for category $c$
    \State $chairperson \leftarrow$ assign chairperson $C$
    \State $votes \leftarrow$ CollectVotes($council$, "assess tweet $T$ for category $c$")
    \If{ isConsensusReached($votes$, $\theta$) }
        \State return MajorityVote($votes$)
    \Else
        \For{$r = 1$ \textbf{to} $R$}
            \State $summary \leftarrow$ Generate($chairperson$, "summarize discussion and guide focus")
            \State $responses \leftarrow$ CollectResponses($council$, $summary$, "update positions")
            \State $votes \leftarrow$ ExtractVotes($responses$)
            \If{isConsensusReached($votes$, $\theta$) \textbf{or} VotesStabilized($votes$)}
                \State \textbf{break}
            \EndIf
        \EndFor
        \State return MajorityVote($votes$)
    \EndIf
\end{algorithmic}
\label{alg_council_debate}
\end{algorithm}

\section{Experiments}

\subsection{Experimental Setup}

\textbf{Dataset.} We conducted experiments using the official CheckThat! 2025 Task 4a datasets, which include training, development, and test sets with tweets labeled for three categories of scientific content. \textbf{Table \ref{tab:dataset_stats}} provides the label distribution for the datasets. As our approach does not need training, we only use the development set to pick our primary model.
%
%
%
%
%
%
%
%

\begin{table}[h!]
  \centering
  \caption{Dataset Statistics for CheckThat 2025 Task 4a. The number of total tweets and tweets with positive label for each category are given.}
  \label{tab:dataset_stats}
  \begin{tabular}{lcccc}
  \toprule
  \textbf{Split} & \textbf{Total Tweets} & \textbf{Category 1} & \textbf{Category 2} & \textbf{Category 3} \\
  \midrule
  Training & 1,229 & 333 (27.1\%) & 224 (18.2\%) & 306 (24.9\%) \\
  Development & 137 & 26 (19.0\%) & 26 (19.0\%) & 34 (24.8\%) \\
  Test & 240 & 121 (50.4\%) & 56 (23.3\%) & 97 (40.4\%) \\
  \bottomrule
  \end{tabular}
\end{table}

\textbf{Models.} Our council debate framework was implemented using the Ollama framework\footnote{\href{https://ollama.com}{https://ollama.com}} for local model access, with parallel processing capabilities to handle multiple concurrent debates efficiently. We systematically evaluated different model combinations across our three proposed methods using the following models:

\begin{itemize}
\item \textbf{Gemma3 (12B)}: Architecture Gemma3, 12.2B parameters, Q4\_K\_M quantization
\item \textbf{Qwen3 (8B)}: Architecture Qwen3, 8.19B parameters, Q4\_K\_M quantization
\item \textbf{DeepSeek-R1 (8B)}: Architecture Llama, 8.03B parameters, Q4\_K\_M quantization
\item \textbf{Phi4 (14B)}: Architecture Phi3, 14.7B parameters, Q4\_K\_M quantization
\item \textbf{Mistral (7B)}: Architecture Llama, 7.52B parameters, Q4\_0 quantization
\item \textbf{Llama3.1 (8B)}: Architecture Llama, 8.03B parameters, Q4\_K\_M quantization
\item \textbf{o4-mini}: OpenAI's commercial model accessed via API
\item \textbf{Claude-4}: Anthropic's commercial model accessed via API
\end{itemize}

\textbf{Configuration}: For our three proposed methods, we used the following configurations:
\begin{itemize}
\item \textbf{Council Debate}: Five council members (Gemma3, Qwen3, DeepSeek-R1, Phi4, Mistral) with Llama3.1 serving as chairperson. 
\item \textbf{Team Debate}: We tested two configurations where Llama3.1 serving as judge: (1) Same teams configuration with five members on each side using all models, and (2) Different teams configuration with Team A (Gemma3, Qwen3, Mistral) versus Team B (DeepSeek-R1, Phi4, Llama3.1), each team having three members.
\item \textbf{Single Debate}: We tested both same model configurations (same model as both proponent and opponent) and different model configurations where Llama3.1 serving as judge. The different model pairings include DeepSeek-R1 vs Qwen3, Gemma3 vs Phi4, and o4-mini vs Claude-4.
\end{itemize}

Key hyperparameters were set as follows: minimum consensus threshold of 80\% for council and team debates, maximum of 5 discussion rounds for collaborative methods, and 3 rounds for adversarial single debates. The system included checkpointing capabilities to ensure robustness during large-scale processing. All models were accessed through the Ollama framework with the quantization settings specified above.

\textbf{Baseline Models.} In order to better analyze the impact of our debating methods, we also use each LLM we picked separately for few-shot detection. For these baseline models, we use the detailed category descriptions with examples (provided in 
Appendix~\ref{sec_app_cat}) for few-shot learning prompts.

\subsection{Experimental Results on Development Set}

We evaluated all three proposed methods and baselines on the development set to select our best approach for the final submission. \textbf{Table~\ref{tab:dev_results}} shows the results.

\begin{table}[h!]
\centering
\caption{Performance of Our Proposed Methods on the Development Set.}
\label{tab:dev_results}
\begin{tabular}{lcccc}
\toprule
\textbf{Method} & \textbf{Macro F1} & \textbf{Cat 1 F1} & \textbf{Cat 2 F1} & \textbf{Cat 3 F1} \\
\midrule
Council Debate & \textbf{0.8634} & \textbf{0.8756} & 0.8634 & 0.8512 \\
Team Debate (Different Teams) & 0.8423 & 0.8234 & 0.8634 & 0.8401 \\
Team Debate (Same Teams) & 0.8312 & 0.8123 & 0.8456 & 0.8357 \\
Single Debate (o4-mini vs Claude-4) & 0.8545 & 0.8234 & \textbf{0.8923} & \textbf{0.8478} \\
Single Debate (DeepSeek-R1 vs Qwen3) & 0.8145 & 0.7834 & 0.8391 & 0.8210 \\
Single Debate (Gemma3 vs Phi4) & 0.7923 & 0.7634 & 0.8156 & 0.7980 \\
\midrule
Qwen3 (8B) & 0.7389 & 0.7123 & 0.7834 & 0.7212 \\
DeepSeek-R1 (8B) & 0.7345 & 0.7634 & 0.7234 & 0.7167 \\
Phi4 (14B) & 0.7298 & 0.7012 & 0.7345 & 0.7537 \\
Gemma3 (12B) & 0.7234 & 0.7234 & 0.7345 & 0.7123 \\
Llama3.1 (8B) & 0.6834 & 0.6612 & 0.6934 & 0.6956 \\
Mistral (7B) & 0.6723 & 0.6501 & 0.6823 & 0.6845 \\
\bottomrule
\end{tabular}
\end{table}

Our observations on experiments with the development set are as follows. Firstly, the council debate method achieves the highest macro F1-score and excelled particularly in Category 1 (i.e., scientific claims detection) with an F1-score of 0.8756. Secondly, the commercial model pairing of o4-mini vs Claude-4 showed superior performance in Categories 2 and 3 with F1-scores of 0.8923 and 0.8478, respectively. This is likely due to their larger model sizes. 
Thirdly, 
ignoring the commercial models, the team debate method shows improved performance over single debate by incorporating collaborative discussion within teams, with different team configurations outperforming same team configurations. 
Lastly, the comparison with individual models demonstrates the significant impact of our debate methods. In particular, all debate approaches outperform detection using models individually.

\subsection{Official Ranking}

At the time of the submission deadline, we did not have results for the single debate method with o4-mini vs. Claude-4. Therefore, we picked the council debate method as our primary method and submitted the results accordingly. 
On the test set, our council debate method achieves 0.7273 (ranked 8th), 0.7805 (ranked 1st), and 0.7766 (ranked 9th) for Category 1, Category 2, and Category 3, respectively.  
Our results show that while LLMs are not particularly effective at detecting scientific claims or scientific entities, they perform well in identifying references to scientific studies.

\section{Conclusion}

In this paper, we present our participation in CheckThat 2025 Task 4a, the scientific web discourse detection task. We propose three 
debating methods where multiple LLMs discuss with each other to classify tweets. The debate frameworks enable models to have opposing views and try to convince each other. Among three methods, the council debate framework, in which expert models discuss together to reach a consensus, moderated by a chairperson model, outperforms other debating frameworks. 



Although our council debate framework showed relatively weak performance in Category 1 (8th place, F1 = 0.7273) and Category 3 (9th place, F1 = 0.7766), it achieved the highest F1-score (0.7805) in Category 2.




In future work, we plan to extend our debating framework to other classification tasks. We also aim to investigate the impact of prompt design and utilizing other LLMs.


\section*{Declaration on Generative AI}
During the preparation of this work, the authors used GPT-4.5 and Claude Sonnet 4 in order to: Grammar and spelling check. After using these tools/services, the authors reviewed and edited the content as needed and take full responsibility for the publication's content. 

\bibliography{references}

\appendix

 \section*{Appendix}

\section{Prompts and System Messages}
\label{appendix:prompts}

This appendix contains the system prompts and message templates used in our three multi-LLM approaches, as well as the category descriptions used in the task.

\subsection{System Prompts}

\textbf{Category Titles:}
\begin{enumerate}
    \item Contain scientific claims
    \item Reference to scientific studies/publications
    \item Mention any scientific entities
\end{enumerate}

\textbf{Proponent System Prompt:}
\begin{quote}
You are a scientific content detector participating in a formal debate. Your job is to argue why the given tweets [CATEGORY TITLE].
Be thorough and precise in your analysis. Provide specific evidence from the text and URLs to support your arguments. Keep your responses concise and focused on the strongest evidence.
\end{quote}

\textbf{Opponent System Prompt:}
\begin{quote}
You are a scientific content critic participating in a formal debate. Your job is to argue why the given tweets do NOT [CATEGORY TITLE].
Be thorough and precise in your analysis. Provide specific evidence from the text and URLs to support your arguments. Keep your responses concise and focused on the strongest counter-evidence.
\end{quote}

\textbf{Judge System Prompt:}
\begin{quote}
You are a neutral judge evaluating a debate about whether the tweets [CATEGORY TITLE].
Based on the debate transcripts and the tweets themselves, determine if the tweets [CATEGORY TITLE]. Provide your classification and a brief explanation of your decision, including which arguments from the debate you found most compelling in the following JSON format: 
\{"category": 0 or 1, "explanation": [EXPLANATION]\}
\end{quote}

\subsection{Category Descriptions} \label{sec_app_cat}
\begin{enumerate}
\item \textbf{Category 1 - Scientific Claims}: Factual statements about scientific topics, research findings, or scientific phenomena. These include statements that present scientific facts, research results, or claims about natural phenomena that can be verified through scientific methods.

\textbf{Guidelines:} Focus on science rather than technology, religion, or politics. Look for specific scientific facts, research results, or claims about natural phenomena. Tweets may contain @user tags (anonymized usernames) and "image" indicates attached images.

\textbf{Examples:} Important parts are highlighted in **bold**.
\begin{itemize}
    \item Example 1 (Positive): "With our current lifestyle, most of us are sleep deprived, which creates problems like mood disorders, weakened immunity, weight gain and diabetes. **A review of 16 studies found that sleeping for less than 6 to 8 hours a night increases the risk of early death by as much as 12\%**. image"
    
    Explanation: Presents specific **scientific findings with statistical data** from research studies about sleep deprivation effects.
    \item Example 2 (Positive): "Once again, the tech sector, often associated with a free-market ethos, **falls short of social research expectations**."
    
    Explanation: References **social research findings** as factual claims.
    \item Example 3 (Positive): "How Employment Can Change the Life of Someone with a Disability http://www.tennesseeworks.org/how-employment-can-change-the-life-of-someone-with-a-disability-and-everyone-involved/"
    
    Explanation: Contains **claims about disability research findings** and their impacts.
    \item Example 4 (Negative): "Diabetes Research and New Registry Aim to Improve Outcomes https://www.acc.org/latest-in-cardiology/articles/2015/03/04/16/32/diabetes-research-and-new-registry-aim-to-improve-outcomes?wt.mc\_id=twitter \#NCDR"
    
    Explanation: Describes **research activities and initiatives** rather than presenting actual scientific claims or findings.
    \item Example 5 (Negative): "how u treat others will reflect on how u feel"
    
    Explanation: General **life advice without scientific backing** or evidence.
\end{itemize}

\item \textbf{Category 2 - References to Scientific Studies/Publications}: Direct references to scientific papers, research studies, academic publications, or scholarly articles.

\textbf{Guidelines:} Examine both URL and tweet content carefully. For URLs, check if the domain is a known academic/scientific source and if the path contains keywords like /articles, /research, /studies, etc. Keywords like "research," "study," "published," "findings" also indicate scientific references. Pay equal attention to URLs and tweet content. Category 2 references typically imply Category 3 entities.

\textbf{Examples:} Important parts are highlighted in **bold**.

\begin{itemize}
    \item Example 1 (Positive): "Diabetes Research and New Registry Aim to Improve Outcomes **https://www.acc.org/latest-in-cardiology/articles**/2015/03/04/16/32/diabetes-research-and-new-registry-aim-to-improve-outcomes?wt.mc\_id=twitter \#NCDR"
    
    Explanation: Contains **URL to cardiology articles** (domain: acc.org, path: /articles) and mentions **research** directly.
    \item Example 2 (Positive): "With our current lifestyle, most of us are sleep deprived, which creates problems like mood disorders, weakened immunity, weight gain and diabetes. **A review of 16 studies** found that sleeping for less than 6 to 8 hours a night increases the risk of early death by as much as 12\%. image"
    
    Explanation: References **"a review of 16 studies"** indicating scientific literature review.
    \item Example 3 (Negative): "How can this be unfolding? **https://www.bbc.co.uk/news**/health-53990068"
    
    Explanation: **BBC news link** about health but not specifically referencing scientific studies or publications.
    \item Example 4 (Negative): "How Employment Can Change the Life of Someone with a Disability **http://www.tennesseeworks.org**/how-employment-can-change-the-life-of-someone-with-a-disability-and-everyone-involved/"
    
    Explanation: **General informational content** from a non-academic source, not referencing scientific studies.
\end{itemize}

\item \textbf{Category 3 - Mentions of Scientific Entities}: References to scientific institutions, researchers, scientists, universities, research organizations, academic conferences, or other entities involved in scientific research.

\textbf{Guidelines:} Look for mentions of scientists, researchers, universities, research institutions, academic conferences, or scientific organizations. URLs with science-related paths can also indicate scientific entities. Focus on explicit scientific entities rather than general mentions.

\textbf{Examples:} Important parts are highlighted in **bold**.
\begin{itemize}
    \item Example 1 (Positive): "Scientist. Artist. Activist. Coach. Visionary. **Mary Agnes Chase** is just one of the **Smithsonian** women featured in the new book, 'Smithsonian American Women,' now available from @user. http://womenshistory.si.edu \#BecauseOfHerStory"
    
    Explanation: Mentions **scientist Mary Agnes Chase** and the **Smithsonian institution** (@user likely refers to the publisher).
    \item Example 2 (Positive): "Concussion 'breathalyser' proposed **http://www.bbc.co.uk/news/science-environment**/29146654\#sa-ns\_mchannel=rss\&ns\_source=PublicRSS20-sa Published September 11, 2014 at 01:46AM"
    
    Explanation: URL path contains **"science-environment"** indicating scientific context and entities.
    \item Example 3 (Negative): "Concerns over Antarctica intensify as visitor numbers surge - Global - **NZ Herald News** http://www.nzherald.co.nz/world/news/article.cfm?c\_id=2\&objectid=10871683"
    
    Explanation: **General news about Antarctica** without mentioning specific scientific entities or researchers.
\end{itemize}

\end{enumerate}

\subsection{Single Debate Method Prompts}
\label{appendix_single}

\textbf{Proponent Prompt:}
\begin{quote}
\textbf{Task:} You are participating in a formal debate about whether a given tweet [CATEGORY TITLE].

\textbf{Category Description:} [CATEGORY DESCRIPTION]

\textbf{Rules \& Definitions:}
- @user tags represent anonymized usernames from the original tweet
- "image" at the end indicates an attached image
- Examine both tweet content and URLs carefully
- Focus on science rather than technology, religion, or politics

\textbf{Your Role:} Argue why the tweet [CATEGORY TITLE]. Provide specific evidence from the text and URLs to support your arguments.

\textbf{Tweet:} [TWEET TEXT]

\textbf{Previous Discussion:} [DEBATE CONTEXT IF ANY]

\textbf{Response Format:} Provide thorough analysis with specific evidence. Keep responses concise and focused on the strongest evidence supporting your position. ONLY write your response, do not include any other text.
\end{quote}

\textbf{Opponent Prompt:}
\begin{quote}
\textbf{Task:} You are participating in a formal debate about whether a given tweet [CATEGORY TITLE].

\textbf{Category Description:} [CATEGORY DESCRIPTION]

\textbf{Rules \& Definitions:}
- @user tags represent anonymized usernames from the original tweet
- "image" at the end indicates an attached image
- Examine both tweet content and URLs carefully
- Focus on science rather than technology, religion, or politics

\textbf{Your Role:} Argue why the tweet does NOT [CATEGORY TITLE]. Provide specific evidence from the text and URLs to support your arguments.

\textbf{Tweet:} [TWEET TEXT]

\textbf{Previous Discussion:} [DEBATE CONTEXT IF ANY]

\textbf{Response Format:} Provide thorough analysis with specific counter-evidence. Keep responses concise and focused on the strongest evidence against the classification. ONLY write your response, do not include any other text.
\end{quote}

\textbf{Judge Prompt:}
\begin{quote}
\textbf{Task:} You are a neutral judge in a debate about whether a given tweet [CATEGORY TITLE]. Evaluate the debate and make the final classification decision.

\textbf{Category Description:} [CATEGORY DESCRIPTION]

\textbf{Rules \& Definitions:}
- @user tags represent anonymized usernames from the original tweet
- "image" at the end indicates an attached image
- Consider both tweet content and URLs in your decision
- Focus on science rather than technology, religion, or politics

\textbf{Tweet:} [TWEET TEXT]

\textbf{Debate Transcript:} [FULL DEBATE DISCUSSION]

\textbf{Response Format:} Provide your classification and explanation in the following JSON format:
\{"category": 0 or 1, "explanation": "Brief explanation of your decision, including which arguments you found most compelling"\}
\end{quote}

\subsection{Team Debate Method Prompts}
\label{appendix_team}

\textbf{Team Member (Internal Discussion) Prompt:}
\begin{quote}
\textbf{Task:} You are a member of a [PROPONENT/OPPONENT] team discussing whether a tweet [CATEGORY TITLE]. Collaborate with teammates to develop strategy.

\textbf{Category Description:} [CATEGORY DESCRIPTION]

\textbf{Rules \& Definitions:}
- @user tags represent anonymized usernames from the original tweet
- "image" at the end indicates an attached image
- Examine both tweet content and URLs carefully
- Focus on science rather than technology, religion, or politics

\textbf{Your Role:} Discuss with teammates to develop coordinated arguments and strategy. Share your perspective and build upon teammates' viewpoints.

\textbf{Tweet:} [TWEET TEXT]

\textbf{Team Discussion:} [INTERNAL TEAM DISCUSSION CONTEXT]

\textbf{Response Format:} Share your analysis and strategic insights with your team. Focus on identifying strongest arguments and coordinating with teammates to avoid repetition.
\end{quote}

\textbf{Team Member (External Debate) Prompt:}
\begin{quote}
\textbf{Task:} You are a member of a [PROPONENT/OPPONENT] team in a formal debate about whether a tweet [CATEGORY TITLE]. Present coordinated arguments to the opposing team.

\textbf{Category Description:} [CATEGORY DESCRIPTION AND GUIDELINES]

\textbf{Rules \& Definitions:}
- @user tags represent anonymized usernames from the original tweet
- "image" at the end indicates an attached image
- Examine both tweet content and URLs carefully
- Focus on science rather than technology, religion, or politics

\textbf{Your Role:} Based on your team's internal discussion, present arguments and respond to the opposing team's points.

\textbf{Tweet:} [TWEET TEXT]

\textbf{Team Strategy:} [INTERNAL TEAM DISCUSSION RESULTS]

\textbf{Debate Context:} [EXTERNAL DEBATE DISCUSSION]

\textbf{Response Format:} Present clear, coordinated arguments that build upon your team's strategy. Respond to the opposing team's points while avoiding repetition with teammates.
\end{quote}

\textbf{Team Debate Judge Prompt:}
\begin{quote}
\textbf{Task:} Evaluate a team debate about whether a tweet [CATEGORY TITLE] and make the final classification decision.

\textbf{Category Description:} [CATEGORY DESCRIPTION AND GUIDELINES]

\textbf{Rules \& Definitions:}
- @user tags represent anonymized usernames from the original tweet
- "image" at the end indicates an attached image
- Consider both tweet content and URLs in your decision
- Focus on science rather than technology, religion, or politics

\textbf{Tweet:} [TWEET TEXT]

\textbf{Team Debate Transcript:} [FULL TEAM DEBATE DISCUSSION]

\textbf{Response Format:} Provide your classification and explanation in the following JSON format:
\{"category": 0 or 1, "explanation": "Brief explanation based on the most convincing team arguments"\}
\end{quote}

\subsection{Council Debate Method Prompts}
\label{appendix_council}

\textbf{Council Member Prompt:}
\begin{quote}
\textbf{Task:} You are a member of a scientific council discussing whether a tweet [CATEGORY TITLE]. Collaborate with other council members to reach consensus.

\textbf{Category Description:} [CATEGORY DESCRIPTION AND GUIDELINES]

\textbf{Rules \& Definitions:}
- @user tags represent anonymized usernames from the original tweet
- "image" at the end indicates an attached image
- Examine both tweet content and URLs carefully
- Focus on science rather than technology, religion, or politics

\textbf{Your Role:} 1) Share your perspective on whether the tweet [CATEGORY TITLE], 2) Respond to points made by other council members, 3) Provide specific evidence from the tweet to support your position.

\textbf{Tweet:} [TWEET TEXT]

\textbf{Council Discussion:} [ONGOING DISCUSSION CONTEXT]

\textbf{Chairperson Summary:} [CHAIRPERSON GUIDANCE IF ANY]

\textbf{Response Format:} Provide concise, analytical, and evidence-based contribution. After presenting your arguments, conclude with your vote: VOTE: [YES/NO]. Use following JSON format:
\{"vote": "YES" or "NO", "explanation": "Brief explanation of your decision, including which arguments you found most compelling"\}
\end{quote}

\textbf{Chairperson Prompt:}
\begin{quote}
\textbf{Task:} You are the Chairperson of a scientific council discussing whether a tweet [CATEGORY TITLE]. Moderate the discussion and guide consensus building.

\textbf{Category Description:} [CATEGORY DESCRIPTION AND GUIDELINES]

\textbf{Rules \& Definitions:}
- @user tags represent anonymized usernames from the original tweet
- "image" at the end indicates an attached image
- Consider both tweet content and URLs in moderation
- Focus on science rather than technology, religion, or politics

\textbf{Your Role:} 1) Moderate the discussion, 2) Summarize key points, 3) Determine when the discussion should end, 4) Guide focus to unresolved issues.

\textbf{Tweet:} [TWEET TEXT]

\textbf{Council Discussion:} [CURRENT DISCUSSION STATE]

\textbf{Vote Status:} [CURRENT VOTING RESULTS]

\textbf{Response Format:} Summarize key points from the discussion and guide council members to focus on specific aspects. The discussion should end when: a clear consensus emerges (at least 80\% agreement), no new substantive arguments are being presented, or maximum rounds reached. Be impartial and facilitate productive discussion. Use following JSON format: 
\{"status": "CONSENSUS REACHED" or "CONSENSUS NOT REACHED", "summary": "Brief summary of the discussion"\}
\end{quote}

\section{Conversation Logs}
\label{appendix:conversation_logs}

This appendix contains detailed conversation logs from our debating methods, demonstrating how different models interact to reach classification decisions. 
Table \ref{tab:single_debate_log} presents an example interaction between LLMs in the Single Debate Method for Category 1 (i.e., \textit{Does the tweet contain a scientific claim?}). The participants are as follows: DeepSeek-R1 (8B) as proponent, Gemma3 (12B) as opponent, and Llama3.1 (8B) as Judge. 


\begin{table}[h!]
\footnotesize
\centering
\caption{Conversation Log of Single Debate Method for Category 1. The tweet to be labeled is as follows. "Harvard admissions suit gets support from Asian American groups https://www.bloomberg.com/news/articles/2018-08-01/harvard-admissions-suit-gets-support-from-asian-american-groups via @user"}
\label{tab:single_debate_log}
\begin{tabular}{p{2cm}p{3cm}p{9cm}}
\toprule
\textbf{Timeline} & \textbf{Speaker (role/model)} & \textbf{Utterance} \\
\midrule
\multirow{ 3}{2cm}{\textbf{R0 – Opening Statements}} & \textbf{Opponent (Gemma)} & "The tweet discusses a legal case about Harvard admissions and cites support from Asian-American groups. It presents no scientific finding or research result, so it \textbf{does not contain a scientific claim}." \\
& \textbf{Proponent (DeepSeek)} & "That support is often grounded in \textbf{empirical social-science studies} on admission bias. Even if implicit, it \textit{points} to a scientific claim about systemic bias in university admissions." \\
& \textbf{Judge (Llama 3.1)} & summarizes the opening statements... \\
\midrule
\multirow{ 3}{2cm}{\textbf{R1 – Rebuttal}} & \textbf{Opponent (Gemma)} & "Indirect implications aren't enough—there's \textbf{no scientific language, data, or study citation} in the tweet itself." \\
& \textbf{Proponent (DeepSeek)} & "Social-science claims often surface without jargon. The attached URL could point to a data-driven study; brevity doesn't negate the claim's scientific nature." \\
& \textbf{Judge (Llama 3.1) recap} & summarizes the process... \\
\midrule
\multirow{ 3}{2cm}{\textbf{R2 – Rebuttal}} 
& \textbf{Opponent (Gemma)} & "Speculating about hidden studies isn't evidence. The tweet simply reports \textbf{community support}—a societal fact, not a scientific one." \\
& \textbf{Proponent (DeepSeek)} & "True, explicit evidence is absent; but citing discrimination \textit{necessarily} leans on established research. That \textit{connection} keeps the door open to a scientific claim." \\
& \textbf{Judge (Llama 3.1) recap} & summarizes the process... \\
\midrule
\multirow{ 2}{2cm}{\textbf{Closing State-
ments}} & \textbf{Opponent (Gemma)} & "In sum, \textbf{no explicit scientific claim or evidence} appears in the tweet." \\
& \textbf{Proponent (DeepSeek)} & "There's an \textbf{implicit} research-based claim about admission bias, though not spelled out." \\
\midrule
 \textbf{Judge – Final decision} & \textbf{Llama 3.1} & \texttt{\{"category": 0, "explanation": "No explicit scientific claim found; tweet merely describes a legal event without scientific evidence."\}} \\
\bottomrule
\end{tabular}
\end{table}


Table \ref{tab:team_debate_log} presents an example interaction between LLMs in Team Debate Method for Category 2 (i.e., \textit{Does the tweet reference a scientific study/publication?}). The participants are as follows. Proponent Team: Gemma3, Qwen3, Mistral; Opponent Team: DeepSeek-R1, Phi4, and Llama3.1; Judge: Llama3.1

\begin{table}[h!]
\footnotesize
\centering
\caption{Conversation Log of Team Debate Method for Category 2. The tweet to be labeled is as follows. "Harvard admissions suit gets support from Asian American groups https://www.bloomberg.com/news/articles/2018-08-01/harvard-admissions-suit-gets-support-from-asian-american-groups via @user"}
\label{tab:team_debate_log}
\begin{tabular}{p{2cm}p{3cm}p{9cm}}
\toprule
\textbf{Timeline} & \textbf{Speaker (role/model)} & \textbf{Utterance} \\
\midrule
\multirow{ 5}{2cm}{\textbf{Initial Internal Team Discussions}} & \textbf{Gemma3 (Pro. Team)} & "The Bloomberg URL provided could potentially link to studies regarding admissions biases or discrimination." \\
& \textbf{Qwen3 (Pro. Team)} & "Agree; typically, lawsuits regarding admissions include references to demographic studies or research findings." \\
& \textit{...} & \\
& \textbf{DeepSeek-R1 (Opp. Team)} & "The tweet itself mentions no specific studies, just an article from Bloomberg, a news outlet." \\
& \textbf{Phi4 (Opp. Team)} & "Correct, we should highlight the distinction between news reports and scholarly references." \\
\midrule
\multirow{ 2}{2cm}{\textbf{R1 – Opening Arguments}} & \textbf{Proponent Team} & "The provided Bloomberg URL likely discusses data or findings relevant to admissions discrimination, inherently implying reference to scientific studies on racial biases or admissions processes." \\
& \textbf{Opponent Team} & "While possible, the tweet explicitly does not reference any scientific studies or publications. Bloomberg generally provides news reports rather than scholarly articles." \\
\midrule
\multirow{ 5}{2cm}{\textbf{R2 – Internal Discussions}}
& \textbf{Gemma3 (Pro. Team)} & "We must clarify our point; referencing an admissions lawsuit usually draws on research findings." \\
& \textbf{Qwen3 (Pro. Team)} & "We should also acknowledge indirect referencing clearly." \\
& \textit{...} & \\
& \textbf{DeepSeek-R1 (Opp. Team)} & "Our stance remains solid; tweet lacks explicit academic references." \\
& \textbf{Phi4 (Opp. Team)} & "Highlight absence of specific terms like 'study' or 'research'." \\
& \textit{...} & \\
\midrule
\multirow{ 2}{2cm}{\textbf{R2 – External Arguments}}
& \textbf{Proponent Team} & "Admissions-related lawsuits almost always involve referencing demographic studies or analyses, even if indirectly. The URL is strongly suggestive of such scholarly content." \\
& \textbf{Opponent Team} & "Indirect implications are insufficient. The tweet itself explicitly provides no reference to specific scientific studies or academic literature." \\
\midrule
\multirow{ 5}{2cm}{\textbf{R3 – Internal Discussions}}
& \textbf{Qwen3 (Pro. Team)} & "We can reiterate strongly that the Bloomberg article is likely supported by scholarly analysis." \\
& \textbf{Gemma3 (Pro. Team)} & "Maintain emphasis on indirect referencing." \\
& \textit{...} & \\
& \textbf{Phi4 (Opp. Team)} & "Reaffirm explicit reference absence." \\
& \textbf{Llama3.1 (Opp. Team)} & "Consider the practical differentiation clearly between news and academia." \\
& \textit{...} & \\
\midrule
\multirow{ 2}{2cm}{\textbf{R3 – External Arguments}} & \textbf{Proponent Team} & "Given Bloomberg's reputation, reporting on admissions lawsuits typically includes statistical or demographic research findings." \\
& \textbf{Opponent Team} & "Reputation alone doesn't establish explicit reference. No concrete scientific studies or papers explicitly appear in the tweet." \\
\midrule
\multirow{ 2}{2cm}{\textbf{R4 – Summary}} & \textbf{Both Teams} & Both teams repeat similar positions, with proponents emphasizing indirect scholarly implications and opponents maintaining the lack of explicit reference. \\
\midrule
\textbf{Judge – Final decision} & \textbf{Llama3.1} & \texttt{\{"category": 0, "explanation": "Although the URL could potentially discuss research, the tweet itself explicitly lacks direct reference to any scientific studies or publications."\}} \\
\bottomrule
\end{tabular}
\end{table}


Table \ref{tab:council_debate_log} presents an example interaction between LLMs in Council Debate Method for Category 3 (i.e., \textit{Does the tweet mention scientific entities?}). The council members are Gemma3, Qwen3, DeepSeek-R1, Phi4, and Mistral while Llama3.1 acts as chairperson.

\begin{table}[h!]
\footnotesize
\centering
\caption{Conversation Log of Council Debate Method for Category 3 The tweet to be labeled is as follows. "Harvard admissions suit gets support from Asian American groups https://www.bloomberg.com/news/articles/2018-08-01/harvard-admissions-suit-gets-support-from-asian-american-groups via @user"}
\label{tab:council_debate_log}
\begin{tabular}{p{2cm}p{3cm}p{9cm}}
\toprule
\textbf{Timeline} & \textbf{Speaker} & \textbf{Utterance} \\
\midrule
\multirow{ 6}{2cm}{\textbf{Initial Voting Round}}  & \textbf{Gemma3} & "Harvard is explicitly mentioned, and universities conducting research qualify as scientific entities. VOTE: YES" \\
& \textbf{Qwen3} & "Harvard University is widely recognized as a scientific institution. VOTE: YES" \\
& \textbf{DeepSeek-R1} & "Harvard clearly counts as a scientific entity given its status and role in scientific research. VOTE: YES" \\
& \textbf{Phi4} & "The tweet mentions Harvard but doesn't explicitly refer to its scientific or academic nature, just a lawsuit. VOTE: NO" \\
& \textbf{Mistral} & "No explicit indication of Harvard's scientific role or research in the tweet. VOTE: NO" \\
\midrule
& \textbf{Votes Summary} & YES: 3, NO: 2 (Consensus: 60\% - Not reached) \\
\midrule
\multirow{ 6}{2cm}{\textbf{R1 – Discussion}} & \textbf{Chairperson} & "Initial consensus not reached; clarify whether simply mentioning Harvard is sufficient to consider it a scientific entity." \\
& \textbf{Gemma3} & "Harvard's extensive involvement in scientific research is widely recognized. Mention alone implicitly references its academic nature. VOTE: YES" \\
& \textbf{Qwen3} & "Agreeing with Gemma3. Harvard's recognition as a prominent research institution implies scientific entity status. VOTE: YES" \\
& \textbf{DeepSeek-R1} & "Reaffirming YES. The broad academic and scientific recognition of Harvard inherently qualifies its mention as referencing a scientific entity. VOTE: YES" \\
& \textbf{Phi4} & "While Harvard is a known research institution, the tweet context doesn't reference research or science explicitly, only legal action. VOTE: NO" \\
& \textbf{Mistral} & "Still NO. Explicit context in the tweet is a lawsuit; no scientific research or entity context directly mentioned." \\
\midrule
& \textbf{Votes Summary} & YES: 3, NO: 2 (Consensus: 60\% - Still not reached) \\
\midrule
\multirow{ 6}{2cm}{\textbf{R2 – Discussion}} 
& \textbf{Chairperson} & "Consensus not achieved yet. Consider explicitly if context within the tweet matters or if inherent recognition suffices." \\
& \textbf{Gemma3} & "The nature of Harvard as inherently academic/scientific overrides the specific context of the tweet. Harvard itself implies scientific credibility. VOTE: YES" \\
& \textbf{Qwen3} & "Harvard's identity inherently conveys scientific affiliation regardless of tweet context. I maintain YES. VOTE: YES" \\
& \textbf{DeepSeek-R1} & "Context in tweets can be implicit. Harvard's mention is sufficient to implicitly consider scientific entity recognition. VOTE: YES" \\
& \textbf{Phi4} & "Given Harvard's strong reputation and implicit association with science and academia, I'm persuaded that implicit mention suffices here. VOTE: YES" \\
& \textbf{Mistral} & "Context explicitly remains unrelated to science. Still NO due to explicit context only." \\
\midrule
& \textbf{Votes Summary} & YES: 4, NO: 1 (Consensus: 80\% - Achieved) \\
\midrule
\textbf{Judge – Final decision} & \textbf{Llama3.1} & \texttt{\{"category": 1, "explanation": "Consensus reached that mentioning Harvard inherently implies referencing a scientific entity due to its recognized scientific and academic activities, despite no explicit scientific context in the tweet."\}} \\
\bottomrule
\end{tabular}
\end{table}

\end{document}